\documentclass{article}

\usepackage{arxiv}

\usepackage[utf8]{inputenc} % allow utf-8 input
\usepackage[T1]{fontenc}    % use 8-bit T1 fonts
\usepackage{hyperref}       % hyperlinks
\usepackage{url}            % simple URL typesetting
\usepackage{booktabs}       % professional-quality tables
\usepackage{amsfonts}       % blackboard math symbols
\usepackage{amssymb}        % \checkmark and other symbols
\usepackage{subcaption}     % subfigures
\usepackage{nicefrac}       % compact symbols for 1/2, etc.
\usepackage{microtype}      % microtypography
\usepackage{graphicx}
\usepackage{natbib}
\usepackage{doi}
\usepackage{amsmath}
\usepackage{listings}
\usepackage{xcolor}
\title{SortScrews: A Dataset and Baseline for Real-time Screw Classification}

\author{ \href{https://orcid.org/0009-0007-2342-5350}{\includegraphics[scale=0.06]{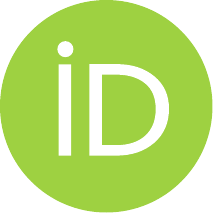}\hspace{1mm}Tianhao Fu}\\
	University of Toronto, Toronto, ON, Canada\\
	Vector Institute, Toronto, ON, Canada\\
	Project Neura, Toronto, ON, Canada\\
	UTMIST, Toronto, ON, Canada\\
	\texttt{terry.fu@projectneura.org} \\
    \And
	{\hspace{1mm}Bingxuan Yang}\\
	University of Toronto, Toronto, ON, Canada\\
    \And
	{\hspace{1mm}Juncheng Guo}\\
	University of Toronto, Toronto, ON, Canada\\
    \And
	{\hspace{1mm}Shrena Sribalan}\\
	University of Toronto, Toronto, ON, Canada\\
	\And
	\href{https://orcid.org/0009-0000-9492-8958}{\includegraphics[scale=0.06]{figures/orcid.pdf}\hspace{1mm}Yucheng Chen}\\
	Project Neura, Toronto, ON, Canada\\
	Amplimit, Toronto, ON, Canada\\
}

\date{}

\renewcommand{\headeright}{Technical Report}
\renewcommand{\undertitle}{Technical Report}
\renewcommand{\shorttitle}{SortScrews}

\hypersetup{
	pdftitle={SortScrews: A Dataset and Baseline for Real-time Screw Classification},
	pdfsubject={cs.CV; cs.AI},
	pdfauthor={Tianhao Fu, Bingxuan Yang, Juncheng Guo, Shrena Sribalan, Yucheng Chen}
}

\begin{document}
	\maketitle
	
    \begin{abstract}
    Automatic identification of screw types is important for industrial automation, robotics, and inventory management. However, publicly available datasets for screw classification are scarce, particularly for controlled single-object scenarios commonly encountered in automated sorting systems. In this work, we introduce \textbf{SortScrews}, a dataset for casewise visual classification of screws. The dataset contains 560 RGB images at $512\times512$ resolution covering six screw types and a background class. Images are captured using a standardized acquisition setup and include mild variations in lighting and camera perspective across four capture settings.
    
    To facilitate reproducible research and dataset expansion, we also provide a reusable data collection script that allows users to easily construct similar datasets for custom hardware components using inexpensive camera setups.
    
    We establish baseline results using transfer learning with EfficientNet-B0 and ResNet-18 classifiers pretrained on ImageNet. In addition, we conduct a well-explored failure analysis. Despite the limited dataset size, these lightweight models achieve strong classification accuracy, demonstrating that controlled acquisition conditions enable effective learning even with relatively small datasets. The dataset, collection pipeline, and baseline training code are publicly available at \url{https://github.com/ATATC/SortScrews}.
    \end{abstract}
	
	\section{Introduction}
    \label{sec:introduction}
    
    \begin{figure}
        \centering
        \includegraphics[width=\linewidth]{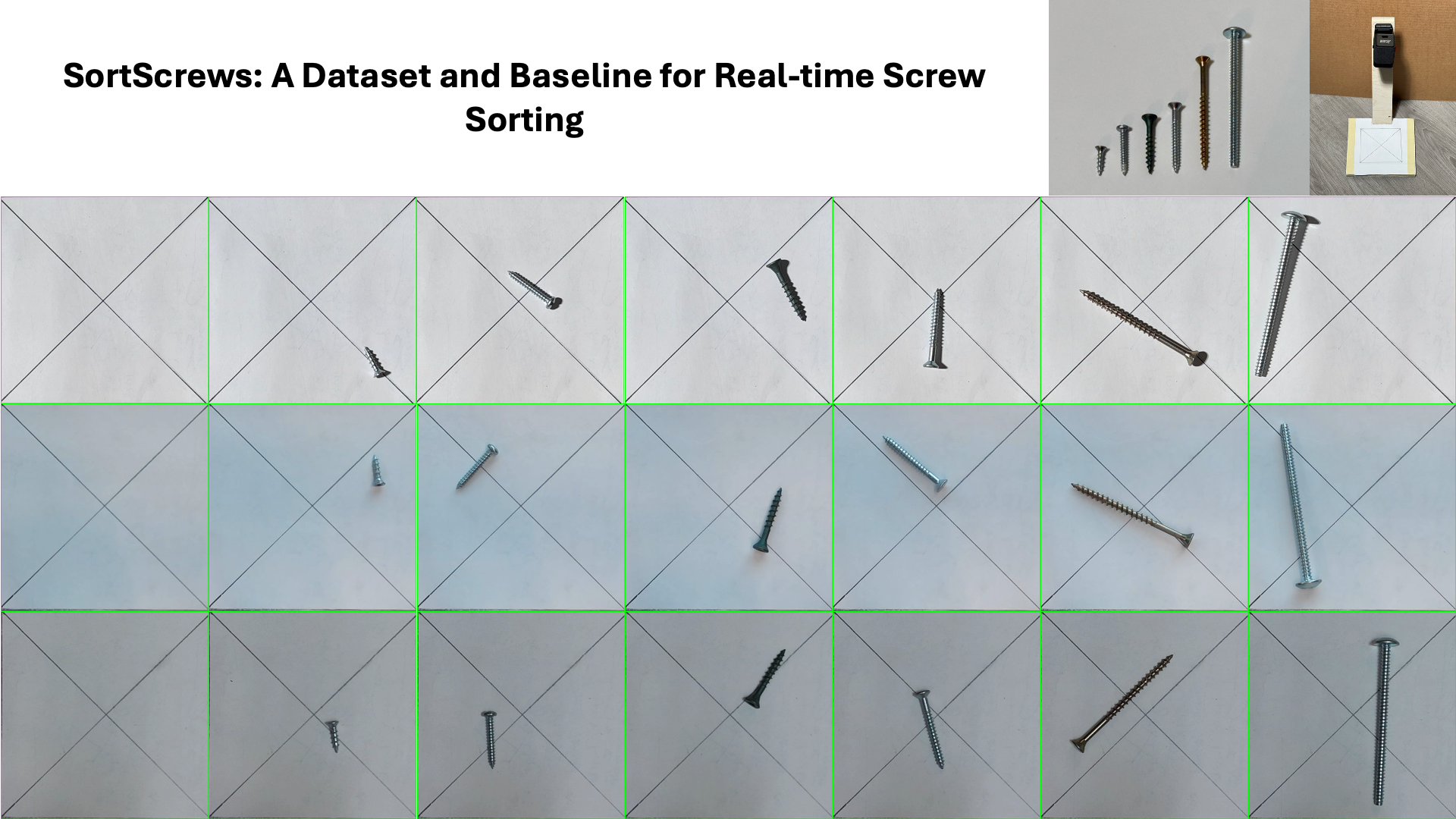}
        \caption{Overview of the SortScrews dataset and acquisition pipeline.}
        \label{fig:poster}
    \end{figure}
    
    Automated recognition of small mechanical components plays an increasingly important role in modern manufacturing and robotics. In particular, screw identification is a fundamental task in automated sorting, inventory management, and robotic assembly systems. Accurate recognition of screw types enables machines to automatically categorize hardware components, streamline manufacturing workflows, and reduce manual inspection costs.
    
    Recent advances in deep learning have significantly improved performance in visual recognition tasks \citep{he2016resnet,tan2019efficientnet}. However, most progress has been driven by large-scale datasets such as ImageNet \citep{deng2009imagenet}, which contain millions of images across thousands of categories. In contrast, datasets for industrial component recognition are relatively scarce, especially for fine-grained object categories such as screws, bolts, and other small hardware parts.
    
    Fine-grained recognition of screws presents several unique challenges. First, screw types often differ only by subtle geometric variations such as head shape, length, or thread pattern. These differences can be difficult for computer vision systems to distinguish without carefully controlled imaging conditions. Second, industrial environments often involve limited labeled data and constrained acquisition setups. As a result, lightweight models and small datasets must be used effectively.
    
    To address these challenges, we introduce \textbf{SortScrews}, a dataset designed for casewise visual classification of screws under controlled acquisition conditions. The dataset consists of 560 RGB images captured using a standardized camera setup. Each image contains a single screw instance placed within a calibrated capture region. The dataset is balanced across six screw categories and includes additional background samples to support rejection in real-world classification pipelines.
    
    In addition to the dataset itself, we provide a reusable data collection pipeline that allows researchers to construct similar datasets using inexpensive camera hardware. The acquisition setup includes a simple physical guide for object placement together with a lightweight data collection script that automatically records labeled samples. This pipeline enables rapid construction of new datasets for custom industrial components or hardware parts.
    
    To establish reference performance on SortScrews, we evaluate two widely used convolutional neural network architectures, EfficientNet-B0 \citep{tan2019efficientnet} and ResNet-18 \citep{he2016resnet}. Both models are trained using transfer learning from ImageNet pretrained weights. Experimental results show that these lightweight models can achieve strong classification accuracy even with relatively small datasets when acquisition conditions are controlled.
    
    Our contributions can be summarized as follows:
    
    \begin{itemize}
    \item We introduce \textbf{SortScrews}, a balanced dataset containing 560 RGB images of six screw types with controlled acquisition conditions.
    \item We release a reusable \textbf{data collection pipeline} that enables rapid creation of similar datasets for industrial components.
    \item We provide \textbf{baseline benchmarks} using EfficientNet-B0 and ResNet-18 to establish reference performance for the dataset.
    \end{itemize}
    
    We release the dataset, data collection scripts, and training pipeline to encourage further research in industrial object recognition and automated sorting systems.
	
	\section{Related Work}
    \label{sec:related}
    
    \subsection{Deep Learning for Image Classification}
    
    Deep convolutional neural networks have achieved remarkable success in image classification tasks. Architectures such as ResNet \citep{he2016resnet} introduced residual connections that enable training of deep networks with improved stability and performance. Later models such as EfficientNet \citep{tan2019efficientnet} further improved parameter efficiency by scaling network depth, width, and resolution in a principled manner.
    
    Large-scale datasets such as ImageNet \citep{deng2009imagenet} have played a crucial role in enabling these advances. Pretraining on large datasets allows models to learn general visual representations that can be transferred to smaller datasets using fine-tuning. This transfer learning paradigm has become standard practice when working with limited training data.
    
    \subsection{Fine-Grained Visual Recognition}
    
    Fine-grained visual classification focuses on distinguishing categories that differ only by subtle visual characteristics. Examples include bird species recognition, car model identification, and aircraft classification. In such tasks, models must identify small geometric differences while remaining robust to variations in lighting, orientation, and background.
    
    Industrial component recognition shares many of these challenges. Objects such as screws or bolts often differ only in small structural details such as head type or thread length. As a result, high-quality image acquisition and consistent object placement are important factors in achieving reliable classification performance.
    
    \subsection{Industrial Vision Datasets}
    
    Several datasets have been proposed for industrial inspection and defect detection. For example, the MVTec AD dataset provides high-resolution images for industrial anomaly detection tasks \citep{bergmann2019mvtec}. These datasets typically focus on identifying defects in manufactured products rather than distinguishing between different types of hardware components.
    
    Datasets specifically targeting small mechanical parts remain relatively limited. In particular, screw classification datasets are rare despite their importance in automated manufacturing pipelines. The lack of standardized datasets makes it difficult to benchmark algorithms for screw recognition and automated sorting.
    
    SortScrews aims to address this gap by providing a small but carefully curated dataset for screw classification together with a reproducible acquisition pipeline. By releasing both the dataset and the data collection tools, we hope to facilitate future research on industrial object recognition and automated sorting systems.
	
	\section{Dataset}

    \subsection{Data Collection Setup}

    \begin{figure}
	    \centering
	    \includegraphics[width=0.25\linewidth]{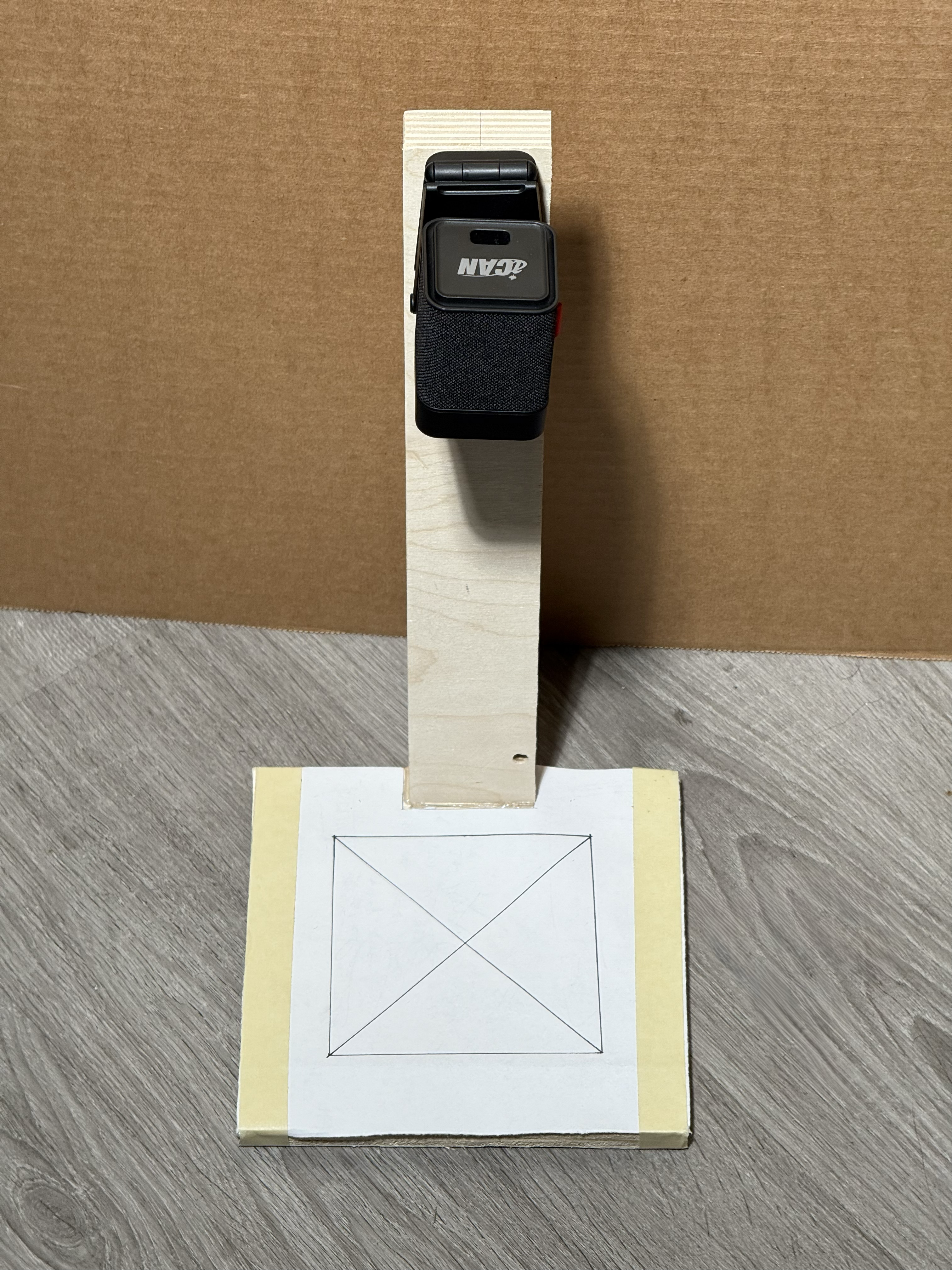}
	    \caption{The device used for data collection, made in Myhal Fabrication Facility.}
	    \label{fig:setup}
	\end{figure}

    Figure~\ref{fig:setup} shows the device we use to collect the dataset. It consists of an iCAN C55N QHD 2K web camera, a wooden stand, and a printed guide for POV calibration.

    We also provide a easy-to-use script for customized dataset collection.

    \subsection{Dataset Overview}

    SortScrews consists of 560 RGB images with a spatial resolution of $512\times512$. The dataset contains six screw categories together with a background class. Each category contains exactly 80 samples, resulting in a balanced dataset.
    
    Images were collected under four acquisition settings that introduce slight variations in lighting conditions and camera perspective. These variations simulate small environmental changes commonly observed in industrial environments. The resulting dataset, therefore, contains both controlled conditions and mild domain variation.
    
    Each image contains a single screw instance placed randomly, indicated by a printed guide on the acquisition surface (Figure~\ref{fig:setup}). The guide ensures consistent object placement while still allowing natural variation in object orientation.
    
    The dataset includes an additional validation set. The validation set contains 28 images sampled uniformly across classes.
    
    \subsection{Screw Categories}

    \begin{figure}
	    \centering
	    \includegraphics[width=0.5\linewidth]{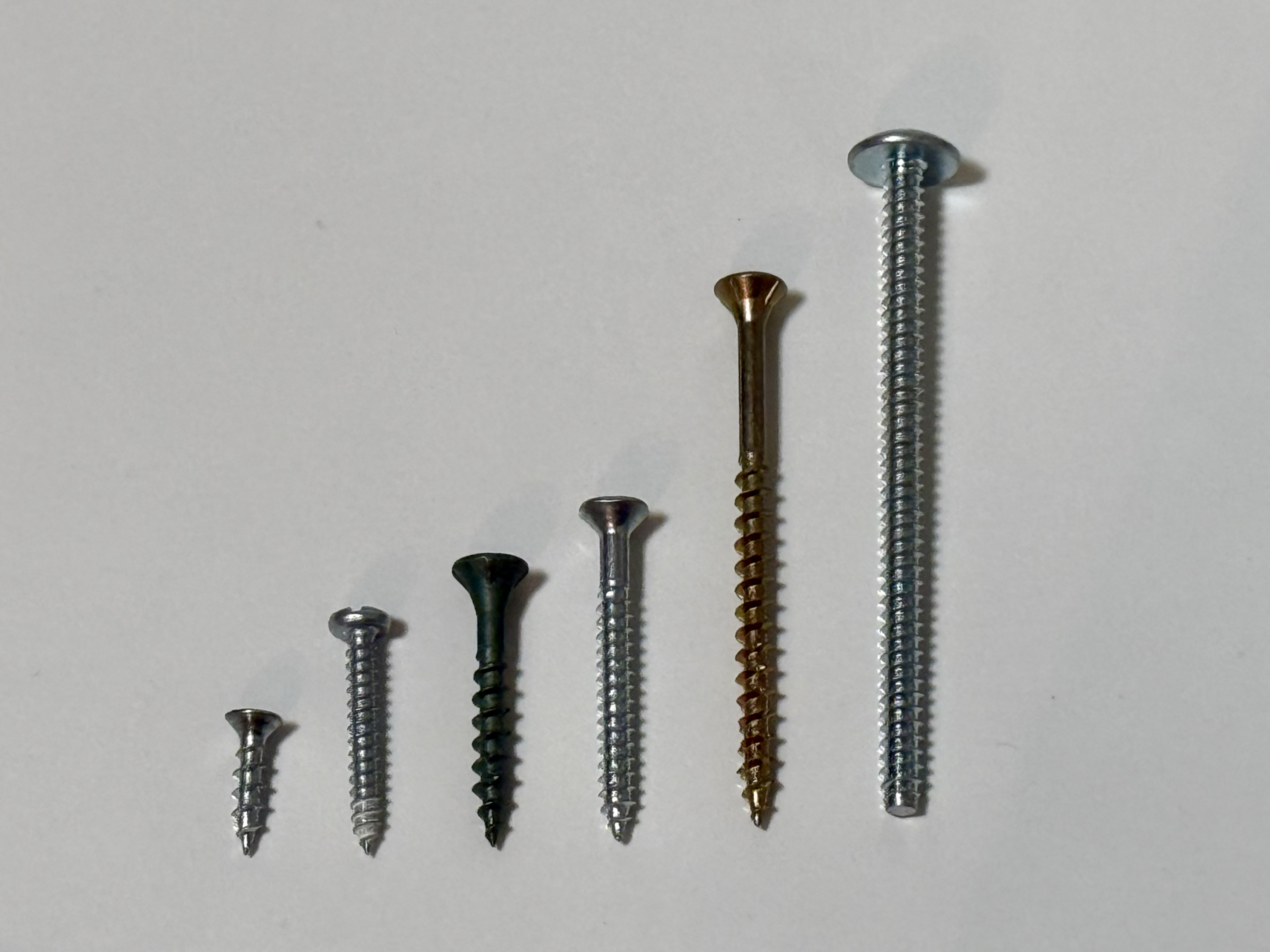}
	    \caption{The six types of screws included in the SortScrews dataset. Their class indices are 1 to 6 from left to right.}
	    \label{fig:screws}
	\end{figure}
    
    The dataset includes six commonly used screw types with varying geometries and materials. The categories were selected to represent visually distinguishable but sometimes subtle differences in screw head design, length, and threading structure.
    
    The screw categories include:
    
    \begin{itemize}
    \item Flat-head 1.5 cm
    \item Round-head 2.5 cm
    \item Flat-head 3.0 cm
    \item Flat-head 3.5 cm
    \item Flat-head 6.0 cm
    \item Round-head 7.5 cm
    \end{itemize}
    
    In addition to these six screw categories, a background class is included to represent images without screws. The background class enables the dataset to support simple rejection mechanisms in real-world sorting pipelines.
    
    Example images for each category are illustrated in Figure~\ref{fig:screws}.
    
    \subsection{Annotation Format}
    
    Annotations are provided in a CSV file containing image filenames and class identifiers. The dataset loader reads this CSV file and associates each image with its corresponding class label.
    
    An example annotation entry is:
    
    \begin{lstlisting}
    case_000.png,0
    \end{lstlisting}
    
    The dataset loader automatically parses this file and returns image tensors together with their class labels. The loader also provides optional filtering to remove background samples if desired. This functionality enables training both binary screw detection models and multi-class screw classifiers. \citep{pytorch2019}
    
    \section{Baseline Methods}
    
    To establish a baseline performance for screw classification, we train a convolutional neural network using transfer learning from a model pretrained on ImageNet \citep{deng2009imagenet}. Transfer learning is commonly used when datasets are relatively small and has been shown to significantly improve performance on fine-grained classification tasks.
    
    \subsection{Network Architectures}

    We evaluate two commonly used convolutional neural network architectures as baselines: EfficientNet-B0 \citep{tan2019efficientnet} and ResNet-18 \citep{he2016resnet}. These models represent two widely adopted design paradigms in modern computer vision.
    
    \paragraph{EfficientNet-B0.}
    EfficientNet models scale network depth, width, and input resolution in a principled manner and achieve strong performance with relatively low computational cost \citep{tan2019efficientnet}. In our experiments, the final classification layer of the pretrained network is replaced with a fully connected layer matching the number of dataset classes.
    
    \paragraph{ResNet-18.}
    ResNet architectures introduce residual connections that enable effective training of deep convolutional networks \citep{he2016resnet}. ResNet-18 is a lightweight model that remains widely used as a strong baseline for image classification tasks. As with EfficientNet, the final classification layer is replaced to match the number of classes in SortScrews.
    
    Both networks are initialized using ImageNet pretrained weights \citep{deng2009imagenet}.
    
    \subsection{Training Procedure}
    
    Training is implemented using \texttt{Trainer} from MIP Candy~\citep{fu2026mipcandymodularpytorch}. Both models are trained using the AdamW optimizer with a learning rate of $10^{-3}$ and weight decay of $10^{-4}$.
    
    Images are resized to $224\times224$ before being fed to the networks.
    
    The training pipeline is summarized as follows:
    
    \begin{itemize}
    \item Batch size: 16
    \item Optimizer: AdamW
    \item Loss: Cross-entropy
    \item Input resolution: $224\times224$
    \item Training epochs: 100
    \end{itemize}
    
    The backbones can optionally and by default be frozen during training to stabilize optimization in small-data regimes. This setting is configurable in the training framework.

    Due to the lightweight nature of our baseline models and the real-time requirement of screw sorting, we train both models on a 2023 14-inch MacBook Pro (Apple M3) with 16 GB unified memory. With Metal acceleration, the epoch time is about 2-3 seconds on average.
    
    \section{Experiments}
    
    \subsection{Evaluation Protocol}
    
    Performance is evaluated using classification accuracy on the validation set. Since the dataset is balanced across classes, accuracy provides an intuitive measure of model performance.
    
    We also report per-class prediction statistics to analyze model behavior and identify systematic classification errors.
    
    \subsection{Results}
    
    Table~\ref{tab:baseline} presents the classification performance of the baseline model.
    
    \begin{table}[h]
    \centering
    \begin{tabular}{lcc}
    \toprule
    Model & Validation Accuracy & Inference Time (ms) \\
    \midrule
    EfficientNet-B0 & 86.2\% & 10.990 ± 4.862 \\
    ResNet-18 & 96.4\% & 19.880 ± 4.083 \\
    \bottomrule
    \end{tabular}
    \caption{Baseline classification performance on the SortScrews dataset.}
    \label{tab:baseline}
    \end{table}
    
    Despite the relatively small dataset size, the baseline models achieve strong performance, demonstrating that visually distinguishable screw types can be learned under controlled acquisition conditions.

    Table~\ref{tab:inference} reports the inference time statistics measured on the full dataset.

    \begin{table}[h]
    \centering
    \begin{tabular}{lcccccc}
    \toprule
    Model & Mean & Std & Median & Min & Max & Throughput \\
    \midrule
    ResNet-18 & 6.42 ms & 1.06 ms & 6.15 ms & 5.09 ms & 10.05 ms & 155.8 fps \\
    EfficientNet-B0 & 17.95 ms & 1.72 ms & 17.92 ms & 14.48 ms & 22.67 ms & 55.7 fps \\
    \bottomrule
    \end{tabular}
    \caption{Single-image inference time on SortScrews (NVIDIA GPU, CUDA).}
    \label{tab:inference}
    \end{table}
    
    \subsection{Failure Analysis}
    
    Although the model performs well overall, some screw types exhibit systematic confusion. In particular, visually similar screw categories, such as screws with similar head shapes or thread patterns, can be misclassified.

    \begin{figure}[h]
    \centering
    \begin{subfigure}{0.48\linewidth}
        \centering
        \includegraphics[width=\linewidth]{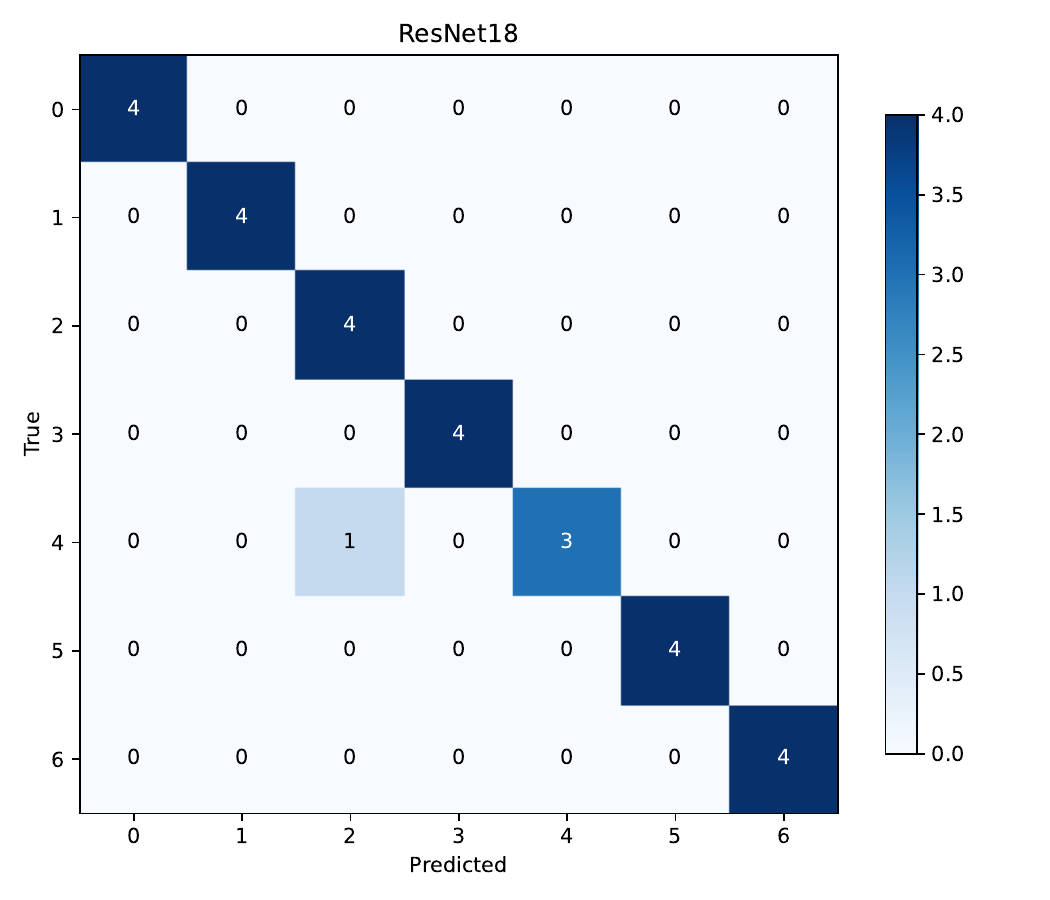}
        \caption{ResNet-18}
    \end{subfigure}
    \hfill
    \begin{subfigure}{0.48\linewidth}
        \centering
        \includegraphics[width=\linewidth]{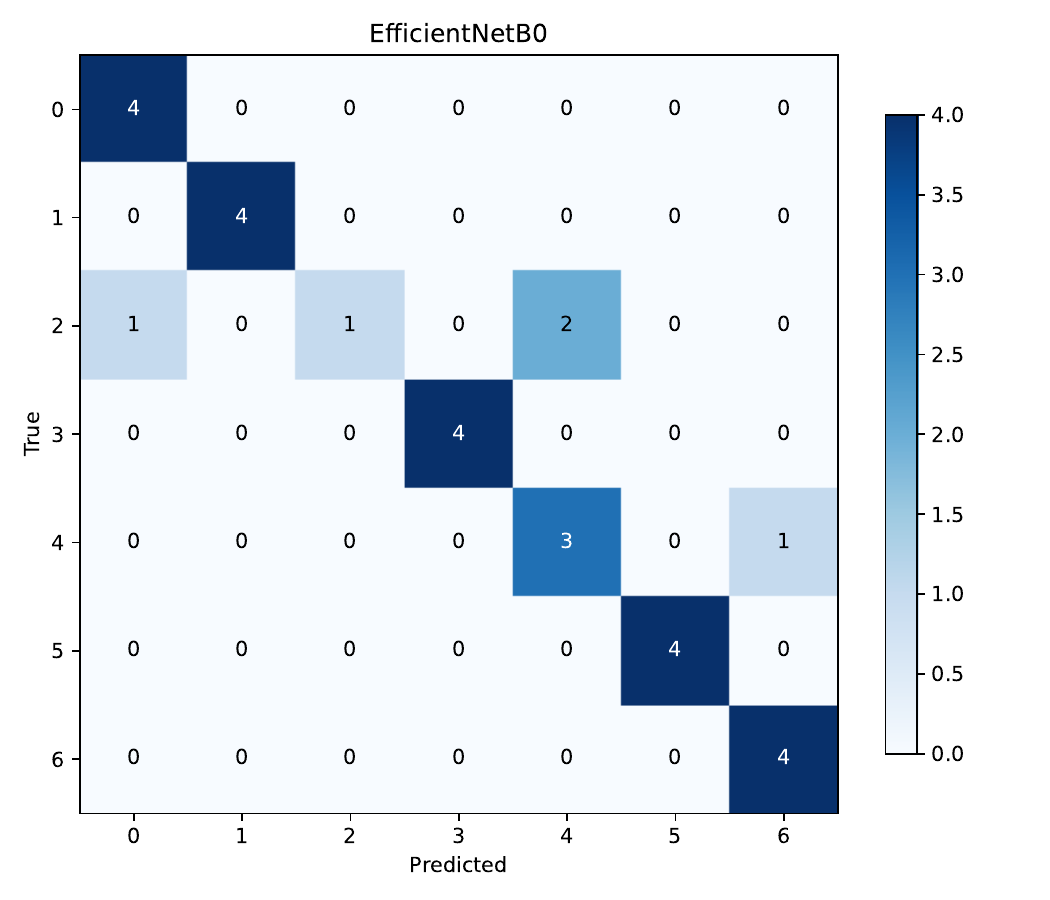}
        \caption{EfficientNet-B0}
    \end{subfigure}
    \caption{Confusion matrices on the validation set.}
    \label{fig:cm}
    \end{figure}

    ResNet-18 misclassifies only one sample (class~4 predicted as class~2), while EfficientNet-B0 exhibits more errors concentrated around classes~2 and~4. In particular, EfficientNet-B0 confuses class~2 (round-head 2.5 cm) with class~4 (flat-head 3.5 cm) and class~0 (background), suggesting difficulty in distinguishing screws with similar lengths but different head shapes.
    
    These errors highlight the challenges of fine-grained industrial object recognition. Subtle geometric differences can be difficult to distinguish without larger datasets or additional viewpoints.

    Furthermore, we noticed that the models learn an unexpected bias in the locations of the screws in the camera view. We believe this is due to the sparse semantic supervision. Explicitly introducing location supervision, as in boundary boxes for object detection, could reduce this sensitivity.
    
    Future work may explore multi-view capture setups, improved lighting conditions, or specialized feature extractors designed for industrial components.
    
    \section{Discussion}
    
    SortScrews demonstrates that useful industrial datasets can be collected using simple and low-cost acquisition setups. The controlled capture environment enables consistent labeling while still introducing moderate variability through lighting and viewpoint changes.
    
    However, the dataset remains relatively small compared to large-scale vision benchmarks. As a result, transfer learning is essential for achieving strong performance.
    
    Future extensions of the dataset could include:
    
    \begin{itemize}
    \item additional screw types
    \item multi-view images
    \item conveyor-belt capture environments
    \item depth or 3D information
    \end{itemize}
    
    Such extensions would further support research in robotic manipulation and automated industrial sorting.

    ResNet-18 consistently outperformed EfficientNet-B0 in our experiments, suggesting that modern architectures do not necessarily improve parameter efficiency nor better capture the subtle visual differences between screw categories.
    
    \section{Conclusion}
    
    We introduced SortScrews, a dataset for screw classification designed to support research in automated sorting and industrial visual recognition. The dataset contains 560 RGB images across six screw categories captured under controlled conditions with mild environmental variation.
    
    We provide baseline results using a transfer learning approach based on EfficientNet-B0. Experimental results demonstrate that accurate classification can be achieved even with a relatively small dataset when acquisition conditions are standardized.
    
    We release the dataset and training pipeline to encourage further research in industrial object recognition and automated sorting systems.
    
    \section*{Acknowledgements}
    
    We thank the Division of Engineering Science, Faculty of Applied Science and Engineering, University of Toronto, for their support in building the data-collection device.
    
    \bibliographystyle{unsrtnat}
	\bibliography{references}

@article{tan2019efficientnet,
  title={EfficientNet: Rethinking model scaling for convolutional neural networks},
  author={Tan, Mingxing and Le, Quoc V},
  journal={International Conference on Machine Learning},
  year={2019}
}

@inproceedings{deng2009imagenet,
  title={ImageNet: A large-scale hierarchical image database},
  author={Deng, Jia and Dong, Wei and Socher, Richard and Li, Li-Jia and Li, Kai and Fei-Fei, Li},
  booktitle={CVPR},
  year={2009}
}

@misc{pytorch2019,
  title={PyTorch: An Imperative Style, High-Performance Deep Learning Library},
  author={Paszke, Adam and others},
  year={2019},
  note={NeurIPS}
}

@article{he2016resnet,
  title={Deep residual learning for image recognition},
  author={He, Kaiming and Zhang, Xiangyu and Ren, Shaoqing and Sun, Jian},
  journal={CVPR},
  year={2016}
}

@misc{fu2026mipcandymodularpytorch,
      title={MIP Candy: A Modular PyTorch Framework for Medical Image Processing}, 
      author={Tianhao Fu and Yucheng Chen},
      year={2026},
      eprint={2602.21033},
      archivePrefix={arXiv},
      primaryClass={cs.CV},
      url={https://arxiv.org/abs/2602.21033}, 
}

@inproceedings{bergmann2019mvtec,
  title={MVTec AD: A Comprehensive Real-World Dataset for Unsupervised Anomaly Detection},
  author={Bergmann, Paul and Fauser, Michael and Sattlegger, David and Steger, Carsten},
  booktitle={Proceedings of the IEEE/CVF Conference on Computer Vision and Pattern Recognition},
  pages={9592--9600},
  year={2019}
}

\end{document}